\pdfoutput=1
\documentclass[conference]{IEEEtran}
\IEEEoverridecommandlockouts
\usepackage{cite}
\usepackage{amsmath,amssymb,amsfonts}
\usepackage{graphicx}
\usepackage{textcomp}
\usepackage{booktabs}
\usepackage[skip=5pt]{caption}
\usepackage{multirow}
\usepackage{hyperref}
\usepackage{url}
\usepackage{nicefrac}
\usepackage{microtype}
\usepackage{xcolor}
\usepackage{algorithm}
\usepackage{algpseudocode}
\usepackage{tikz}
\usepackage{quantikz}
\usepackage{orcidlink}
\usepackage{comment}
\usepackage{braket}
\usepackage{stfloats}
\usepackage{placeins}  

\graphicspath{{./}}

\def\BibTeX{{\rm B\kern-.05em{\sc i\kern-.025em b}\kern-.08em
    T\kern-.1667em\lower.7ex\hbox{E}\kern-.125emX}}

\begin{document}
\bstctlcite{bstctl:etal, bstctl:nodash, bstctl:simpurl}

\title{QFT-Based Hybrid PointNet-Quantum Neural Networks for
       LiDAR Point-Cloud Classification}


\author{\IEEEauthorblockN{Anurag Kulkarni\textsuperscript{1}, Nouhaila Innan\textsuperscript{2,3},
        and Muhammad Shafique\textsuperscript{2,3}}
\IEEEauthorblockA{
\textsuperscript{1}Deep Tech R\&D Lab, Anukul Consultancy Services (AKCS), Maharashtra, India\\
\textsuperscript{2}eBRAIN Lab, Division of Engineering,
  New York University Abu Dhabi (NYUAD), Abu Dhabi, UAE\\
\textsuperscript{3}Center for Quantum and Topological Systems (CQTS),
  NYUAD Research Institute, NYUAD, Abu Dhabi, UAE\\
anukul984@gmail.com, nouhaila.innan@nyu.edu, muhammad.shafique@nyu.edu}}

\maketitle

\begin{abstract}
Hybrid quantum--classical learning architectures provide a promising framework for integrating quantum feature transformations into complex data-processing tasks. Among these transformations, the Quantum Fourier Transform (QFT) offers a structured approach to feature processing, motivating an investigation of how its circuit configuration influences predictive performance, quantum resource requirements, and noise robustness. We present a hybrid PointNet--quantum neural network architecture for LiDAR point-cloud classification, combining classical feature extraction, amplitude encoding, and a variational quantum circuit. Five Fourier-processing configurations are systematically evaluated: No-QFT, Full-QFT-After, QFT-no-swap, IQFT-After, and Trainable-QFT. Experiments on KITTI, nuScenes, and PandaSet across three random seeds assess classification performance, class-level behavior, circuit complexity, and sensitivity to the IBM FakeManilaV2 noise model. Under ideal simulation, IQFT-After achieves the highest weighted F1-score on KITTI ($0.857\pm0.033$), compared with $0.838\pm0.023$ for No-QFT. Removing the final SWAP gates in the QFT circuit reduces circuit complexity while preserving the Pauli-$Z$ measurement features up to a fixed permutation that can be absorbed by the classification head. Under quantum noise, the circuit ranking changes, with IQFT-After exhibiting greater performance degradation and No-QFT and Trainable-QFT showing stronger robustness. Per-class analysis further identifies dataset-dependent differences in minority-class performance. These results demonstrate that Fourier-processing configuration influences both classification behavior and noise sensitivity, highlighting the role of circuit structure and hardware-aware evaluation in hybrid quantum learning for three-dimensional point-cloud data.
\end{abstract}


\section{Introduction}
\label{sec:introduction}

LiDAR-based point-cloud classification plays an important role in autonomous driving, robotics, and three-dimensional (3D) scene understanding \cite{abbasi2022lidar,geiger2012kitti}. By capturing the spatial distribution of objects and their surrounding environments, LiDAR sensors provide geometric information for perception and decision-making systems. However, learning from point-cloud data presents several challenges, including irregular spatial distributions, varying point density, measurement noise, and imbalanced object classes. Classical deep learning architectures, particularly PointNet and its extensions \cite{qi2017pointnet,qi2017pointnetpp}, have demonstrated strong capabilities in learning geometric representations from unordered point sets. These challenges motivate the study of alternative feature-processing mechanisms that can be integrated into existing point-cloud learning architectures.

Quantum machine learning (QML) introduces quantum information processing into learning models through quantum state encoding, parameterized transformations, and measurement-based feature extraction \cite{schuld2019qml,cerezo2021vqa,benedetti2019pqc}. Hybrid quantum--classical architectures combine classical feature extraction with variational quantum circuits (VQCs), enabling quantum operations to process compact learned representations \cite{mari2020transfer}. The design of these architectures involves considerations of quantum resource requirements, circuit complexity, and noise sensitivity, which influence their execution on current quantum hardware and their scalability toward fault-tolerant systems. The behavior of hybrid quantum models depends on several design choices, including data encoding, variational circuit structure, entanglement patterns, and measurement observables. Understanding how these components influence learning performance and quantum resource requirements is an important step toward developing hardware-aware quantum learning systems.

Among quantum transformations, the Quantum Fourier Transform (QFT) provides a structured mapping between computational and Fourier bases. It is a fundamental component of quantum algorithms, including quantum phase estimation and factoring \cite{nielsen2010,shor1997}. Beyond these applications, Fourier-based representations have attracted increasing attention in quantum machine learning, with recent studies examining the relationships between data encoding, circuit structure, and the frequency characteristics of quantum models \cite{schuld2021fourier,strobl2026fourier,oh2026fourier}. The use of QFT-based interference for learning and inference has also been investigated, providing further motivation for incorporating Fourier transformations into quantum learning architectures \cite{wakeham2024inference}. Within hybrid quantum models, an explicit QFT operator transforms the learned quantum state before measurement, influencing the information extracted through quantum observables.

The effectiveness of QFT-based feature processing depends on the structure of the transformation and its interaction with the surrounding variational circuit. Different configurations, including forward QFT, inverse QFT, and parameterized controlled-phase transformations, introduce distinct operations and phase relationships into the quantum feature-processing stage. Moreover, the final SWAP gates in the conventional QFT circuit reverse the qubit ordering, which raises questions about their necessity when quantum features are extracted through expectation-value measurements and subsequently processed by a trainable classical classifier. These architectural choices also influence circuit depth, gate requirements, and exposure to hardware noise. We systematically compare these configurations to characterize how QFT configurations affect classification performance, measurement outcomes, and quantum execution costs.

Motivated by these considerations, we develop a hybrid PointNet--quantum neural network framework for LiDAR point-cloud classification and investigate the impact of Fourier-based quantum feature processing. The framework combines permutation-invariant geometric feature extraction, classical feature compression, amplitude encoding, variational quantum processing, and measurement-based classification. We examine five configurations: No-QFT, Full-QFT-After, QFT-no-swap, IQFT-After, and Trainable-QFT. The evaluation covers three LiDAR datasets, KITTI, nuScenes, and PandaSet, under consistent experimental conditions. Beyond classification performance, we investigate class-level behavior, learned feature representations, circuit resource requirements, and sensitivity to quantum noise.

The main contributions of this work are summarized as follows:

\begin{itemize}
    \item 
    We develop a PointNet-based architecture integrating classical geometric feature extraction with variational quantum processing and configurable Fourier transformations for LiDAR point-cloud classification.

    \item 
    We examine five circuit configurations under matched training conditions, characterizing the influence of Fourier-transform direction, final SWAP operations, and trainable controlled-phase rotations on quantum feature processing and classification performance.

    \item 
    We conduct experiments on KITTI, nuScenes, and PandaSet across three random seeds, evaluating classification accuracy, weighted F1-score, learned feature representations, and minority-class behavior.

    \item 
    We characterize circuit complexity through gate counts, circuit depth, and trainable quantum parameters, and evaluate the sensitivity of each configuration using the IBM FakeManilaV2 noise model.

    \item 
    We analyze the trade-offs between ideal-simulation performance, circuit structure, and noise robustness, including the removal of final QFT SWAP operations for expectation-value readout followed by a trainable classification head.
\end{itemize}

\section{Background and Related Work}
\label{sec:related_work}

\subsection{Point-Cloud Learning for LiDAR Data}

LiDAR point clouds represent three-dimensional environments through unordered and irregularly distributed spatial measurements, presenting challenges for geometric feature extraction and classification. PointNet \cite{qi2017pointnet} established a permutation-invariant learning framework based on shared point-wise transformations and symmetric global pooling, while PointNet++ \cite{qi2017pointnetpp} introduced hierarchical neighborhood aggregation to capture local geometric structures. Subsequent methods explored alternative representations, including voxel-based feature extraction in VoxelNet \cite{zhou2018voxelnet}, sparse convolutional processing in SECOND \cite{yan2018second}, and pillar-based representations in PointPillars \cite{lang2019pointpillars}. These architectures established fundamental approaches for learning from irregular point-cloud data.

More recent research has focused on improving geometric representation learning and computational efficiency. VoxelNeXt \cite{chen2023voxelnext} introduced a fully sparse detection framework that processes voxel features without dense prediction heads. SAFDNet \cite{zhang2024safdnet} extended sparse feature processing through adaptive feature diffusion to improve long-range 3D object detection. Transformer-based architectures have also gained attention for modeling spatial relationships within point clouds. Point Transformer V3 \cite{wu2024point} employs serialized point-cloud representations to improve processing efficiency and expand the effective receptive field, demonstrating strong performance across indoor and outdoor perception tasks. More recently, UniMamba \cite{jin2025unimamba} combined spatial convolution and state-space modeling to capture local and global dependencies for LiDAR-based 3D object detection. These developments illustrate the importance of feature representation, neighborhood modeling, and computational requirements in modern point-cloud learning.

Beyond architectural design, robustness and generalization remain important considerations for LiDAR perception. Eskandar \cite{eskandar2024empirical} systematically evaluated the effects of network architecture, voxel encoding, augmentation, and detection strategies under sensor, weather, and geographic domain shifts. Mali\'c et al. \cite{malic2025gblobs} introduced explicit local geometric representations to improve cross-domain LiDAR detection. Class imbalance presents an additional challenge, particularly for less frequently observed object categories, motivating approaches such as class-weighted objectives and focal loss \cite{lin2017focal}. In object-level classification, these factors influence the quality and consistency of learned geometric representations across different datasets.

Multimodal approaches have further investigated the integration of geometric and visual information through LiDAR--camera feature fusion \cite{chen2017mv3d}. Kulkarni \cite{kulkarni2025lidar} developed a multimodal PointNet-based architecture combining LiDAR and RGB information for four-class KITTI classification, reporting 87\% validation accuracy on a selected dataset subset. Building on the PointNet feature-extraction approach, our work investigates the integration of variational quantum circuits and Fourier-based quantum transformations for object-level point-cloud classification. The focus is on characterizing how quantum feature-processing configurations influence classification performance, circuit resource requirements, and noise sensitivity across multiple LiDAR datasets.

\subsection{Hybrid Quantum Neural Networks}

Hybrid quantum--classical neural networks integrate classical feature extraction with parameterized quantum circuits for learning from encoded representations. Their performance is influenced by data encoding, variational circuit structure, entanglement, and measurement selection, which determine the accessible feature representations and affect trainability and generalization \cite{innan2024financial,kashif2026design,el2026comparative}. Circuit architecture also influences hardware execution through gate complexity, connectivity requirements, and sensitivity to quantum noise.

Among encoding strategies, amplitude encoding maps a normalized $2^n$-dimensional feature vector onto $n$ qubits, although arbitrary state preparation can introduce substantial gate costs. Recent research has examined alternative encoding mechanisms and their resource requirements \cite{kyriacou2026shot,innan2026spate}. These considerations motivate systematic evaluations of quantum circuit architectures that account for both predictive performance and execution cost. In this work, we investigate how Fourier-based transformations modify the measured representations and resource requirements of a hybrid quantum classifier.

\subsection{Quantum Fourier Transform in QML}

The QFT is a fundamental quantum operation that maps computational-basis states into Fourier-basis representations through Hadamard gates, controlled-phase rotations, and qubit-reordering operations \cite{nielsen2010}. Beyond its established role in quantum algorithms, its structured transformations have motivated research on Fourier-based quantum learning and feature representations.

The Fourier properties of parameterized quantum models have been extensively investigated in relation to data encoding and expressivity. Schuld et al. \cite{schuld2021fourier} established the Fourier-series representation of quantum models with data-dependent encoding gates, while Strobl et al. \cite{strobl2026fourier} characterized how variational circuit architectures constrain correlations among Fourier coefficients. Recent analyses have further examined the relationships between encoding Hamiltonians, accessible frequency components, and quantum model behavior \cite{oh2026fourier,novaes2026fourier}. These findings provide a theoretical foundation for analyzing the representational structure of quantum learning models.

Beyond Fourier-series analysis, explicit QFT operations have been investigated as components of quantum learning methods. Wakeham and Schuld \cite{wakeham2024inference} examined Fourier-space interference for learning and inference, using the identification of invariant structures as a guiding principle. Payares and Mart\'inez \cite{payares2023enhancement} integrated QFT operations into quantum machine learning models and evaluated their performance under ideal simulation and IBM quantum hardware execution. Their results demonstrated the potential of QFT-based circuit modifications for improving learning performance. These studies establish different approaches to incorporating Fourier transformations into quantum learning architectures.

While Fourier-series analysis characterizes the input-dependent functions represented by quantum models, an explicit QFT operator transforms the encoded quantum state before measurement. Its influence on classification depends on the surrounding variational circuit, the Fourier-processing configuration, and the measurement observables. Building on existing QFT-based learning approaches, this work systematically investigates forward, inverse, swap-free, and trainable QFT configurations under matched experimental conditions. The analysis characterizes their effects on classification performance, quantum resource requirements, and noise robustness across multiple LiDAR datasets.

\section{Problem Formulation}
\label{sec:problem_formulation}

Consider a supervised LiDAR point-cloud classification dataset
$\mathcal{D}=\{(\mathcal{P}_i,y_i)\}_{i=1}^{M}$, where
$\mathcal{P}_i=\{\mathbf{p}_{i,j}\}_{j=1}^{N}$ denotes a point cloud (an annotated object region for nuScenes and
PandaSet, a full scan for KITTI) containing $N$ three-dimensional points, with $\mathbf{p}_{i,j}\in\mathbb{R}^{3}$, and
$y_i\in\mathcal{Y}$ is its corresponding class label.
The classification task considers four object categories,
$\mathcal{Y}=\{\mathrm{Car},\mathrm{Pedestrian},
\mathrm{Cyclist},\mathrm{DontCare}\}$. Following preprocessing,
each point cloud is represented by
$\bar{\mathcal{P}}_i\in\mathbb{R}^{N\times3}$.

The objective is to learn a hybrid quantum--classical
classification function
\begin{equation}
f_{\boldsymbol{\Theta},v}:
\mathbb{R}^{N\times3}\rightarrow[0,1]^C,
\qquad
\sum_{c=1}^{C}\hat{p}_c^{(v)}=1,
\label{eq:classification_mapping}
\end{equation}
where $C=|\mathcal{Y}|=4$, $\boldsymbol{\Theta}$ denotes
the trainable classical and quantum parameters, and
$v\in\mathcal{V}$ identifies the Fourier-processing
configuration. For an input point cloud
$\bar{\mathcal{P}}$, the model produces a probability
vector $\hat{\mathbf{p}}^{(v)}
=f_{\boldsymbol{\Theta},v}(\bar{\mathcal{P}})$.

The hybrid model first extracts a classical
representation $\mathbf{h}$, which is compressed and
encoded into a quantum state. A parameterized quantum
circuit transforms the encoded state, followed by a
configuration-dependent Fourier operator $F_v$.
Quantum measurements are subsequently combined with
a classical feature descriptor to generate the final
classification output. The resulting mapping is
expressed as
\begin{equation}
\hat{\mathbf{p}}^{(v)}
=
\mathcal{C}_{\boldsymbol{\omega}}
\left(
\mathcal{M}\left[
F_v U_{\mathrm{VQC}}(\boldsymbol{\theta})
|\psi_{\mathrm{in}}\rangle
\right],
H(\mathbf{h})
\right),
\label{eq:hybrid_mapping}
\end{equation}
where $U_{\mathrm{VQC}}(\boldsymbol{\theta})$ denotes
the variational quantum circuit, $\mathcal{M}$ extracts
quantum expectation values, $H(\mathbf{h})$ is a
classical entropy descriptor, and
$\mathcal{C}_{\boldsymbol{\omega}}$ represents the
classification head.

We characterize the influence of $F_v$ on classification performance, quantum circuit complexity, and noise sensitivity.
Five configurations are investigated: No-QFT,
Full-QFT-After, QFT-no-swap, IQFT-After, and
Trainable-QFT. All configurations use the same
classical feature extractor, encoding strategy,
variational circuit structure, measurement
observables, and classification architecture.
This controlled formulation enables the effects
of Fourier-processing structure to be evaluated
under consistent architectural conditions.
\section{Methodology}
\label{sec:methodology}

\begin{figure*}[t]
    \centering
    \includegraphics[width=\linewidth]{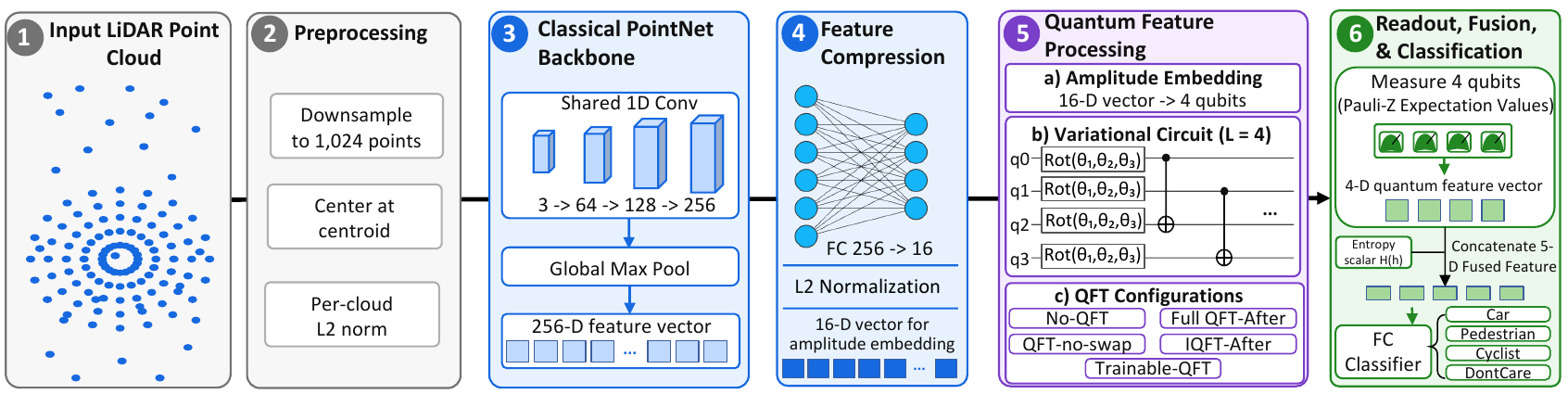}
    \caption{Overview of the hybrid quantum--classical framework for LiDAR point-cloud classification. Input point clouds are resampled to 1,024 points, centered, and normalized. A PointNet backbone extracts a 256-dimensional feature representation, which is compressed to 16 dimensions for four-qubit amplitude encoding. A four-layer variational circuit is followed by a configurable Fourier-processing stage. Pauli-$Z$ expectation values from the four qubits are concatenated with a Shannon-entropy descriptor derived from the PointNet features. The resulting five-dimensional representation is passed to a fully connected classifier for four-class prediction.}
    \label{fig:methodology}
\end{figure*}

The proposed hybrid quantum--classical framework integrates PointNet-based feature extraction with variational quantum processing for LiDAR point-cloud classification. As illustrated in Fig.~\ref{fig:methodology}, the architecture comprises six stages: (1) input LiDAR point clouds, (2) point-cloud preprocessing, (3) permutation-invariant feature extraction using a PointNet backbone, (4) classical feature compression, (5) quantum feature processing through amplitude encoding, variational layers, and a configuration-dependent Fourier transformation, and (6) quantum readout, feature fusion, and classification.

To investigate the influence of Fourier processing, we evaluate five circuit configurations: No-QFT, Full-QFT-After, QFT-no-swap, IQFT-After, and Trainable-QFT. The classical backbone, feature compression, amplitude encoding, variational ansatz, measurement observables, and classification architecture remain consistent across configurations. The comparison isolates the effects of Fourier transformation direction, final qubit reordering, and trainable phase rotations on classification performance, circuit resources, and noise sensitivity.

\subsection{Input LiDAR Point Cloud and Preprocessing}
\label{subsec:preprocessing}

Each input LiDAR point cloud (an object region or, for KITTI, a full scan) is resampled to a fixed size of $N=1{,}024$ points. For an input containing $M$ points, random sampling without replacement is applied when $M\geq N$, while sampling with replacement is used when $M<N$.

The resampled point cloud is centered at its centroid and normalized by the maximum Euclidean distance from the centroid:
\begin{equation}
\boldsymbol{\mu}
=
\frac{1}{N}\sum_{i=1}^{N}\mathbf{p}_i,
\qquad
\bar{\mathbf{p}}_i
=
\frac{\mathbf{p}_i-\boldsymbol{\mu}}
{\max_{1\leq j\leq N}\|\mathbf{p}_j-\boldsymbol{\mu}\|_2+\epsilon},
\label{eq:point_preprocessing}
\end{equation}
where $\mathbf{p}_i\in\mathbb{R}^{3}$ denotes a sampled point, $\boldsymbol{\mu}$ is the point-cloud centroid, and $\epsilon>0$ is a small constant for numerical stability.

The resulting point cloud $\bar{\mathcal{P}}\in\mathbb{R}^{N\times3}$ is invariant to global translation and normalized with respect to object scale, while retaining the relative spatial arrangement of its points.

\subsection{Classical PointNet Backbone}
\label{subsec:pointnet}

A PointNet-style backbone is employed to extract permutation-invariant geometric features from the normalized LiDAR point clouds. The backbone consists of three shared point-wise convolutional layers with channel dimensions $3\rightarrow64\rightarrow128\rightarrow256$, each followed by a ReLU activation. For each point $\bar{\mathbf{p}}_i$, the shared feature extractor $g_{\boldsymbol{\phi}}$ produces a representation $\mathbf{u}_i=g_{\boldsymbol{\phi}}(\bar{\mathbf{p}}_i)\in\mathbb{R}^{256}$, where $\boldsymbol{\phi}$ denotes the trainable backbone parameters.

The point-wise features are aggregated through channel-wise global max pooling:
\begin{equation}
h_j=\max_{1\leq i\leq N}
\left[g_{\boldsymbol{\phi}}(\bar{\mathbf{p}}_i)\right]_j,
\qquad j=1,\ldots,256,
\label{eq:pointnet_pooling}
\end{equation}
yielding a global representation $\mathbf{h}\in\mathbb{R}^{256}$. The shared point-wise mapping and symmetric pooling operation ensure that $\mathbf{h}$ is invariant to permutations of the input point ordering. This representation serves as the input to the subsequent feature-compression stage.

\subsection{Feature Compression}
\label{subsec:compression}

The 256-dimensional PointNet representation is compressed into a lower-dimensional latent space compatible with quantum amplitude encoding. For a circuit with $n_q$ qubits, the input dimension is $d_q=2^{n_q}$. Accordingly, the four-qubit architecture requires a 16-dimensional feature vector. A trainable fully connected layer performs this projection:
\begin{equation}
    \mathbf{a}
    =
    W_{\mathrm{cmp}}\mathbf{h}
    +
    \mathbf{b}_{\mathrm{cmp}},
    \label{eq:feature_compression}
\end{equation}
where $W_{\mathrm{cmp}}\in\mathbb{R}^{16\times256}$, $\mathbf{b}_{\mathrm{cmp}}\in\mathbb{R}^{16}$, and $\mathbf{a}\in\mathbb{R}^{16}$. The compression layer is optimized jointly with the quantum circuit and classification head.

Before quantum state preparation, the compressed features are normalized to unit Euclidean norm:
\begin{equation}
    \mathbf{x}
    =
    \frac{\mathbf{a}}{\|\mathbf{a}\|_2},
    \qquad
    \|\mathbf{x}\|_2^2
    =
    \sum_{j=0}^{15}|x_j|^2
    =
    1,
    \label{eq:feature_normalization}
\end{equation}
for $\mathbf{a}\neq\mathbf{0}$. The resulting vector $\mathbf{x}\in\mathbb{R}^{16}$ provides the input to the amplitude-encoding stage.

\subsection{Quantum Feature Processing}
\label{subsec:quantum_processing}

The quantum feature-processing module applies a
parameterized transformation to the amplitude-encoded
PointNet representation, followed by a configurable
Fourier operator. The architecture comprises $n_q=4$
qubits and $L=4$ variational layers. Across the five
evaluated configurations, the encoding procedure and
variational ansatz are unchanged, allowing the
influence of the Fourier-processing operator to be
examined independently of other architectural choices.

\subsubsection{Amplitude Encoding}

The normalized latent representation
$\mathbf{x}\in\mathbb{R}^{16}$ is mapped to a
four-qubit pure state through amplitude encoding:
\begin{equation}
|\psi_{\mathrm{in}}(\mathbf{x})\rangle
=
\sum_{j=0}^{2^{n_q}-1}x_j|j\rangle,
\qquad
\sum_{j=0}^{2^{n_q}-1}|x_j|^2=1,
\label{eq:quantum_encoding}
\end{equation}
where $|j\rangle$ denotes the $j$th
computational-basis state. The encoded amplitudes
preserve the normalized latent coefficients, while
the subsequent quantum operations determine how
these coefficients contribute to the measured
representation.

Amplitude encoding requires state preparation, whose gate cost depends on the input state and preparation method. Since the
same encoding is used for all configurations,
the Fourier-stage resource comparison isolates
the additional operations introduced after
state preparation.

\subsubsection{Variational Quantum Circuit}

The encoded state is transformed by a
parameterized unitary
$U_{\mathrm{VQC}}(\boldsymbol{\theta})$:
\begin{equation}
|\psi_{\mathrm{var}}\rangle
=
U_{\mathrm{VQC}}(\boldsymbol{\theta})
|\psi_{\mathrm{in}}(\mathbf{x})\rangle,
\label{eq:variational_state}
\end{equation}
where $\boldsymbol{\theta}$ contains the
trainable rotation parameters.

The variational ansatz consists of $L$ layers
of single-qubit rotations and entangling
operations. In each layer $\ell$, a
three-parameter Euler rotation is applied
to qubit $q$:
\begin{equation}
R_{\ell,q}
=
\operatorname{Rot}
\left(
\theta_{\ell,q}^{(1)},
\theta_{\ell,q}^{(2)},
\theta_{\ell,q}^{(3)}
\right),
\label{eq:vqc_rotation}
\end{equation}
with $\ell\in\{1,\ldots,L\}$ and
$q\in\{0,\ldots,n_q-1\}$.

The rotation layers are interleaved with
CNOT-based entangling operations. For the
cyclic connectivity specified in the
architecture, the entangling block is
\begin{equation}
E=
\operatorname{CNOT}_{3,0}
\operatorname{CNOT}_{2,3}
\operatorname{CNOT}_{1,2}
\operatorname{CNOT}_{0,1},
\label{eq:vqc_entangling}
\end{equation}
where the rightmost operation acts first.
Defining
$R_\ell=\bigotimes_{q=0}^{n_q-1}R_{\ell,q}$,
the corresponding variational unitary is
\begin{equation}
U_{\mathrm{VQC}}(\boldsymbol{\theta})
=
(ER_L)\cdots(ER_2)(ER_1).
\label{eq:vqc_unitary}
\end{equation}

Each rotation contributes three independent
parameters per qubit. The total number of
trainable parameters in the variational block is
\begin{equation}
N_{\mathrm{VQC}}
=
3Ln_q=3(4)(4)=48.
\label{eq:vqc_parameters}
\end{equation}
The variational unitary provides a common
trainable transformation for all evaluated
Fourier configurations.

\subsubsection{QFT Configurations}

A Fourier-processing operator $F_v$ is
applied after the variational circuit:
\begin{equation}
|\psi_v(\mathbf{x})\rangle
=
F_v U_{\mathrm{VQC}}(\boldsymbol{\theta})
|\psi_{\mathrm{in}}(\mathbf{x})\rangle,
\label{eq:fourier_processing}
\end{equation}
where $v$ identifies the circuit configuration.
The Fourier operator acts on the variational
output state before measurement, modifying
the observables' expectation values through
a structured unitary transformation.

The $n_q$-qubit QFT is defined as
\begin{equation}
F|j\rangle
=
\frac{1}{\sqrt{2^{n_q}}}
\sum_{k=0}^{2^{n_q}-1}
e^{2\pi\mathrm{i}jk/2^{n_q}}|k\rangle.
\label{eq:qft_operator}
\end{equation}
Its conventional decomposition uses Hadamard
gates, controlled-phase rotations, and final
SWAP gates that reverse the qubit ordering.
For $n_q=4$, the full QFT comprises four
Hadamard gates, six controlled-phase gates,
and two SWAP gates.

The five configurations investigate the
presence, direction, qubit reordering, and
phase parameterization of the Fourier
transformation.

\paragraph{No-QFT}
The reference configuration omits Fourier
processing, $F_{\mathrm{No\text{-}QFT}}=I$.
Measurements are obtained directly from
the variational output state, providing
a reference for assessing the effects of
the additional Fourier operator.

\paragraph{Full-QFT-After}
The forward transform is applied after the
variational circuit,
$F_{\mathrm{Full}}=F$.
This configuration evaluates the effect of
the QFT, including its final
qubit-reversal operations, on the measured
representation.

\paragraph{QFT-no-swap}
The QFT core is retained, while the final
SWAP gates are omitted. Let $S$ denote
the qubit-reversal operator and
$F_{\mathrm{ns}}$ the swap-free circuit.
The operators satisfy
\begin{equation}
F=SF_{\mathrm{ns}},
\qquad
F_{\mathrm{ns}}=S^\dagger F.
\label{eq:swap_free_relation}
\end{equation}
The removal of $S$ eliminates two logical
SWAP gates in the four-qubit implementation.
For Pauli-$Z$ expectation-value readout,
the resulting qubit permutation can be
absorbed by the trainable linear
classification head.

\paragraph{IQFT-After}
The inverse QFT is applied,
$F_{\mathrm{IQFT}}=F^\dagger$.
Its implementation reverses the forward-QFT
gate sequence and conjugates the
controlled-phase rotations. This
configuration examines the effect of
Fourier transformation direction on
the measured quantum features.

\paragraph{Trainable-QFT}
The QFT gate topology is preserved, while
the six fixed controlled-phase angles are
replaced by trainable parameters.
Each controlled-phase gate is defined as
\begin{equation}
\operatorname{CP}(\varphi_{q,r})
=
\operatorname{diag}
(1,1,1,e^{\mathrm{i}\varphi_{q,r}}),
\label{eq:trainable_controlled_phase}
\end{equation}
where $\varphi_{q,r}$ is associated with
a controlled-phase interaction between
qubits $q$ and $r$.

The resulting Fourier operator is denoted
by $F(\boldsymbol{\varphi})$, with
$\boldsymbol{\varphi}\in\mathbb{R}^{6}$.
Its phase parameters are initialized to
the corresponding canonical QFT angles,
\begin{equation}
\boldsymbol{\varphi}^{(0)}
=
\left(
\frac{\pi}{2},
\frac{\pi}{4},
\frac{\pi}{8},
\frac{\pi}{2},
\frac{\pi}{4},
\frac{\pi}{2}
\right)^{\top},
\label{eq:trainable_qft_initialization}
\end{equation}
such that
$F(\boldsymbol{\varphi}^{(0)})=F$,
provided the gate ordering matches the
canonical decomposition.

With six additional phase parameters, the
Trainable-QFT configuration contains
\begin{equation}
N_{\mathrm{Trainable\text{-}QFT}}
=
3Ln_q+\binom{n_q}{2}
=
54
\label{eq:trainable_quantum_parameters}
\end{equation}
trainable quantum parameters.

The trainable phases allow the circuit to
depart from the canonical Fourier
transformation during optimization,
enabling comparison of
fixed Fourier processing and
task-adapted phase transformations.

\paragraph{SWAP-Free Readout Equivalence}
The final SWAP gates of the QFT implement a fixed permutation of the qubit ordering. Let $S$ denote this permutation and $F_{\mathrm{ns}}$ the QFT operator without the final SWAP gates, such that $F=SF_{\mathrm{ns}}$. For Pauli-$Z$ expectation-value measurements, the corresponding quantum feature vectors satisfy
\begin{equation}
\mathbf{z}^{(\mathrm{Full})}
=
P\mathbf{z}^{(\mathrm{no\text{-}swap})},
\end{equation}
where $P$ denotes the qubit-reversal permutation matrix.

Since the subsequent classification layer is trainable and the classical entropy descriptor remains unchanged, this permutation can be absorbed into the classifier weights. Consequently, Full-QFT and QFT-no-swap have equivalent classification function classes under ideal execution. Differences in experimental performance may arise from optimization and noise effects, while the removal of the final SWAP gates reduces the logical circuit operations.

\subsection{Readout, Fusion, and Classification}
\label{subsec:readout_fusion_classification}

The output of each Fourier-processing configuration is
mapped to classical features through single-qubit
Pauli-$Z$ expectation values:
\begin{equation}
    z_q^{(v)}
    =
    \langle\psi_v|Z_q|\psi_v\rangle,
    \qquad q=0,\ldots,n_q-1,
    \label{eq:quantum_readout}
\end{equation}
where $Z_q$ acts on qubit $q$ and as the identity
on all other qubits. The resulting quantum feature
vector $\mathbf{z}^{(v)}\in[-1,1]^{n_q}$ contains
four expectation values that characterize the
transformed state through the selected measurement
observables.

To incorporate information from the classical
representation, a scalar Shannon-entropy descriptor
is computed from the PointNet features
$\mathbf{h}\in\mathbb{R}^{256}$. The feature
activations are normalized into a probability
distribution:
\begin{equation}
    \rho_j
    =
    \frac{\exp(h_j)}
    {\sum_{k=1}^{256}\exp(h_k)},
    \qquad j=1,\ldots,256,
    \label{eq:feature_distribution}
\end{equation}
and the corresponding entropy is
\begin{equation}
    H(\mathbf{h})
    =
    -\sum_{j=1}^{256}\rho_j\log\rho_j.
    \label{eq:feature_entropy}
\end{equation}
The entropy summarizes the concentration of
activations across feature channels without
introducing additional trainable parameters.

The quantum measurements and classical entropy
descriptor are combined into a five-dimensional
representation:
\begin{equation}
    \mathbf{r}^{(v)}
    =
    \begin{bmatrix}
        \mathbf{z}^{(v)}\\
        H(\mathbf{h})
    \end{bmatrix}
    \in\mathbb{R}^{n_q+1}.
    \label{eq:fused_representation}
\end{equation}

A fully connected classification layer maps
the fused features to class probabilities:
\begin{equation}
    \hat{\mathbf{p}}^{(v)}
    =
    \operatorname{softmax}
    \left(
        W_{\mathrm{cls}}\mathbf{r}^{(v)}
        +\mathbf{b}_{\mathrm{cls}}
    \right),
    \label{eq:classification_output}
\end{equation}
where
$W_{\mathrm{cls}}\in\mathbb{R}^{C\times(n_q+1)}$
and $\mathbf{b}_{\mathrm{cls}}\in\mathbb{R}^{C}$.
The predicted label is
$\hat{y}^{(v)}=\arg\max_c\hat{p}_c^{(v)}$.

The same readout observables, fusion procedure,
and classifier architecture are used across
all five Fourier configurations. Consequently,
the measured features provide a consistent
basis for examining the influence of the
Fourier-processing operator on classification.

\section{Experimental Setup}
\label{sec:experimental_setup}

The experimental evaluation is designed to assess the classification performance, noise robustness, and quantum resource requirements of the five Fourier-processing configurations under consistent conditions. This section details the datasets, training procedure, simulation environments, and evaluation protocols, with the main experimental settings summarized in Table~\ref{tab:experimental_configuration}.

\subsection{Datasets and Data Preparation}
\label{subsec:datasets}

The proposed framework is evaluated using three LiDAR benchmarks: KITTI~\cite{geiger2012kitti}, nuScenes~\cite{caesar2020nuscenes}, and PandaSet~\cite{xiao2021pandaset}. The datasets contain point-cloud observations acquired under different sensing configurations and environmental conditions. The evaluation considers four unified classification categories: Car, Pedestrian, Cyclist, and DontCare.

For nuScenes and PandaSet, annotated object regions are extracted using the corresponding three-dimensional bounding boxes. Dataset-specific semantic labels are mapped to the four target categories, with vehicle and pedestrian subclasses consolidated according to predefined mappings. Categories outside the selected groups are assigned to DontCare. The resulting point clouds are resampled to 1,024 points and normalized as described in Section~\ref{subsec:preprocessing}. For KITTI,
each of the 7,481 training scans is resampled to 1,024 points and assigned a
single label from its annotation file (the category of the first annotated
object, with categories outside the selected groups mapped to DontCare).

The datasets exhibit imbalanced class distributions. In the reported KITTI corpus, Cars account for 75.3\% of labeled samples (5,632 instances), whereas Cyclists account for 2.3\% (173 instances), corresponding to a Car-to-Cyclist ratio of approximately 32.6:1. To examine classification behavior across categories, per-class F1 scores are reported alongside aggregate metrics.

For model development, each dataset is randomly divided into 80\% training and 20\% validation samples using a seed-controlled split. The same split-generation procedure is applied across circuit configurations to support consistent comparisons.

\subsection{Implementation and Training Configuration}
\label{subsec:training_configuration}

All hybrid configurations are trained end-to-end, jointly optimizing the PointNet backbone, classical feature-compression layer, variational quantum circuit, and classification head. Trainable-QFT additionally optimizes the six controlled-phase parameters of the Fourier operator. The classical Shannon-entropy descriptor is computed from the softmax-normalized PointNet activations, following Section~\ref{subsec:readout_fusion_classification}, and introduces no additional trainable parameters.

The training objective is standard multiclass cross-entropy. For a mini-batch of $B$ samples, the loss is defined as
\begin{equation}
    \mathcal{L}_{\mathrm{CE}}
    =
    -\frac{1}{B}
    \sum_{b=1}^{B}
    \log \hat{p}_{b,y_b},
    \label{eq:training_loss}
\end{equation}
where $\hat{p}_{b,y_b}$ is the predicted probability assigned to the ground-truth class $y_b$. No class-dependent loss weighting is applied.

All configurations use the Adam optimizer with a learning rate of $10^{-3}$, a mini-batch size of 8, and a maximum of 80 epochs. Early stopping monitors validation weighted F1 with a patience of eight epochs. The checkpoint achieving the highest validation weighted F1 is retained for evaluation.

Each configuration is trained using three random seeds, $\mathcal{S}=\{7,42,123\}$, controlling model initialization and data partitioning. Results are reported as the mean and standard deviation across the three runs. Classification performance is assessed using accuracy and weighted F1, with per-class F1 scores included to characterize differences across the
target categories. Classification average precision (AP) is additionally
computed for each class from the softmax scores, as the area under the
one-vs-rest precision--recall curve, and averaged over Car, Pedestrian, and
Cyclist where a mean is reported. Per-class F1 and AP are evaluated on a fixed
class-stratified subset of 500 samples per dataset, shared across
configurations and seeds. The fixed subset is sampled independently of the seed-specific
train--validation partitions and may include training samples.
Its per-class F1 and AP scores characterize performance on
this shared subset.

\subsection{Quantum Simulation, Noise Modeling, and Resource Evaluation}
\label{subsec:quantum_evaluation}

Ideal quantum experiments are conducted using PennyLane's \texttt{default.qubit} state-vector simulator. Single-qubit Pauli-$Z$ expectation values are evaluated analytically, and gradients are propagated through the differentiable quantum circuits during end-to-end training.

To investigate sensitivity to hardware-derived noise, trained checkpoints are evaluated using IBM's \texttt{FakeManilaV2} fake backend through the PennyLane--Qiskit interface. This backend provides a simulated execution environment informed by stored device calibration characteristics, including gate and measurement errors. Noise evaluation is restricted to inference, with no additional parameter optimization.

For each dataset, a class-stratified evaluation subset of approximately 200 samples is constructed using a fixed sampling seed. Each trained checkpoint is evaluated on the same subset under ideal and noisy execution. Noisy expectation values are estimated using 1,024 measurement shots per circuit evaluation. Noise sensitivity is quantified by the differences in accuracy and weighted F1 between the two execution settings.

Quantum resources are characterized at both logical and backend-transpiled levels. Logical analysis measures circuit depth, total gate count, and two-qubit operation count for the variational and Fourier-processing stages. Single-qubit Euler rotations are expanded into their constituent rotations, while CNOT, controlled-phase, Hadamard, and SWAP operations are retained as logical gates.

For hardware-targeted resource estimation, representative circuits are constructed in Qiskit and transpiled to the \texttt{FakeManilaV2} backend target using optimization levels 0 and 3 and a fixed transpiler seed of 42. The resulting circuits are characterized by their compiled depth, total gate count, and native two-qubit gate count, incorporating gate decomposition, connectivity constraints, and routing overhead.

Resource analysis is performed both with and without amplitude-state preparation. The former estimates the execution requirements of a representative quantum circuit, while the latter isolates the variational and Fourier stages. Representative nonzero rotation parameters are used for compilation, with fixed canonical Fourier angles for the standard QFT variants and specified noncanonical phase values for Trainable-QFT. Accordingly, the transpiled measurements characterize the selected circuit instances and compilation settings, instead of every possible trained parameter configuration.

\begin{table}[t]
\centering
\caption{Experimental configuration.}
\label{tab:experimental_configuration}
\begin{tabular}{ll}
\hline
\textbf{Parameter} & \textbf{Setting} \\
\hline
Datasets & KITTI, nuScenes, PandaSet \\
Classification categories & 4 \\
Points per cloud & 1,024 \\
PointNet feature dimension & 256 \\
Compressed feature dimension & 16 \\
Number of qubits & 4 \\
Variational layers & 4 \\
Quantum trainable parameters & 48 / 54 \\
Optimizer & Adam \\
Learning rate & $10^{-3}$ \\
Loss function & Cross-entropy \\
Batch size & 8 \\
Maximum epochs & 80 \\
Early stopping patience & 8 \\
Random seeds & 7, 42, 123 \\
Ideal simulator & PennyLane \texttt{default.qubit} \\
Noisy backend & IBM \texttt{FakeManilaV2} \\
Noise-evaluation subset & Approximately 200 samples \\
Measurement shots & 1,024 \\
Transpilation levels & 0 and 3 \\
Transpiler seed & 42 \\
\hline
\end{tabular}
\end{table}

\section{Results and Discussion}

\subsection{Classification Performance and Convergence}
\label{subsec:classification_performance}

The classification performance of the five circuit configurations is evaluated across KITTI, nuScenes, and PandaSet to examine the influence of Fourier processing on predictive accuracy and training behavior. The comparison considers weighted F1 and classification accuracy under ideal quantum simulation, with results averaged over three independent training seeds. Table~\ref{tab:simulated_3ds} summarizes the performance across datasets, highlighting differences in classification outcomes and variability among circuit configurations.

On KITTI, IQFT-After achieves the highest mean weighted F1
($0.857\pm0.033$) and accuracy ($0.873\pm0.025$), improving upon
No-QFT by 1.9 and 1.5 percentage points, respectively. Full-QFT-After
and Trainable-QFT perform comparably to the baseline, whereas
QFT-no-swap obtains the lowest mean F1 ($0.803\pm0.063$) and
the largest variation across seeds. These results indicate
differences in both classification performance and training
variability among the Fourier configurations.

On nuScenes, the configurations achieve closely grouped weighted
F1 scores ($0.901$--$0.906$), with Full-QFT-After and IQFT-After
sharing the highest mean. This narrow range indicates that the
additional Fourier transformations yield limited differences in
classification performance on this dataset. PandaSet exhibits
a different ranking, with Trainable-QFT attaining the highest
weighted F1 ($0.836\pm0.018$), a 1.7-percentage-point improvement
over No-QFT, followed by IQFT-After ($0.830\pm0.027$).
The variation in rankings across datasets indicates that the
effectiveness of Fourier processing depends on the underlying
classification setting, with no configuration consistently
leading across all three benchmarks.

\begin{table}[!t]
    \centering
    \caption{Classification performance under ideal quantum simulation
    across three datasets (mean $\pm$ standard deviation over three
    seeds).}
    \label{tab:simulated_3ds}
    \setlength{\tabcolsep}{4pt}
    \renewcommand{\arraystretch}{1.1}
    \small
    \begin{tabular}{clcc}
        \toprule
        & \textbf{Configuration}
        & \textbf{Weighted F1}
        & \textbf{Accuracy} \\
        \midrule

        \multirow{5}{*}{\rotatebox[origin=c]{90}{\textit{KITTI}}}
        & No-QFT         & $0.838\pm0.023$ & $0.858\pm0.017$ \\
        & Full-QFT-After & $0.837\pm0.018$ & $0.857\pm0.016$ \\
        & QFT-no-swap    & $0.803\pm0.063$ & $0.837\pm0.034$ \\
        & IQFT-After     & $\mathbf{0.857\pm0.033}$ & $\mathbf{0.873\pm0.025}$ \\
        & Trainable-QFT  & $0.836\pm0.020$ & $0.857\pm0.012$ \\

        \midrule

        \multirow{5}{*}{\rotatebox[origin=c]{90}{\textit{nuScenes}}}
        & No-QFT         & $0.902\pm0.001$ & $0.904\pm0.002$ \\
        & Full-QFT-After & $\mathbf{0.906\pm0.008}$ & $0.907\pm0.007$ \\
        & QFT-no-swap    & $0.901\pm0.005$ & $0.904\pm0.003$ \\
        & IQFT-After     & $\mathbf{0.906\pm0.010}$ & $\mathbf{0.908\pm0.008}$ \\
        & Trainable-QFT  & $0.905\pm0.009$ & $0.906\pm0.010$ \\

        \midrule

        \multirow{5}{*}{\rotatebox[origin=c]{90}{\textit{PandaSet}}}
        & No-QFT         & $0.819\pm0.021$ & $0.846\pm0.009$ \\
        & Full-QFT-After & $0.816\pm0.018$ & $0.845\pm0.007$ \\
        & QFT-no-swap    & $0.829\pm0.028$ & $0.850\pm0.013$ \\
        & IQFT-After     & $0.830\pm0.027$ & $0.848\pm0.011$ \\
        & Trainable-QFT  & $\mathbf{0.836\pm0.018}$ & $\mathbf{0.852\pm0.011}$ \\

        \bottomrule
    \end{tabular}
\end{table}

\begin{figure*}[ht]
    \centering
    \includegraphics[width=\linewidth]
    {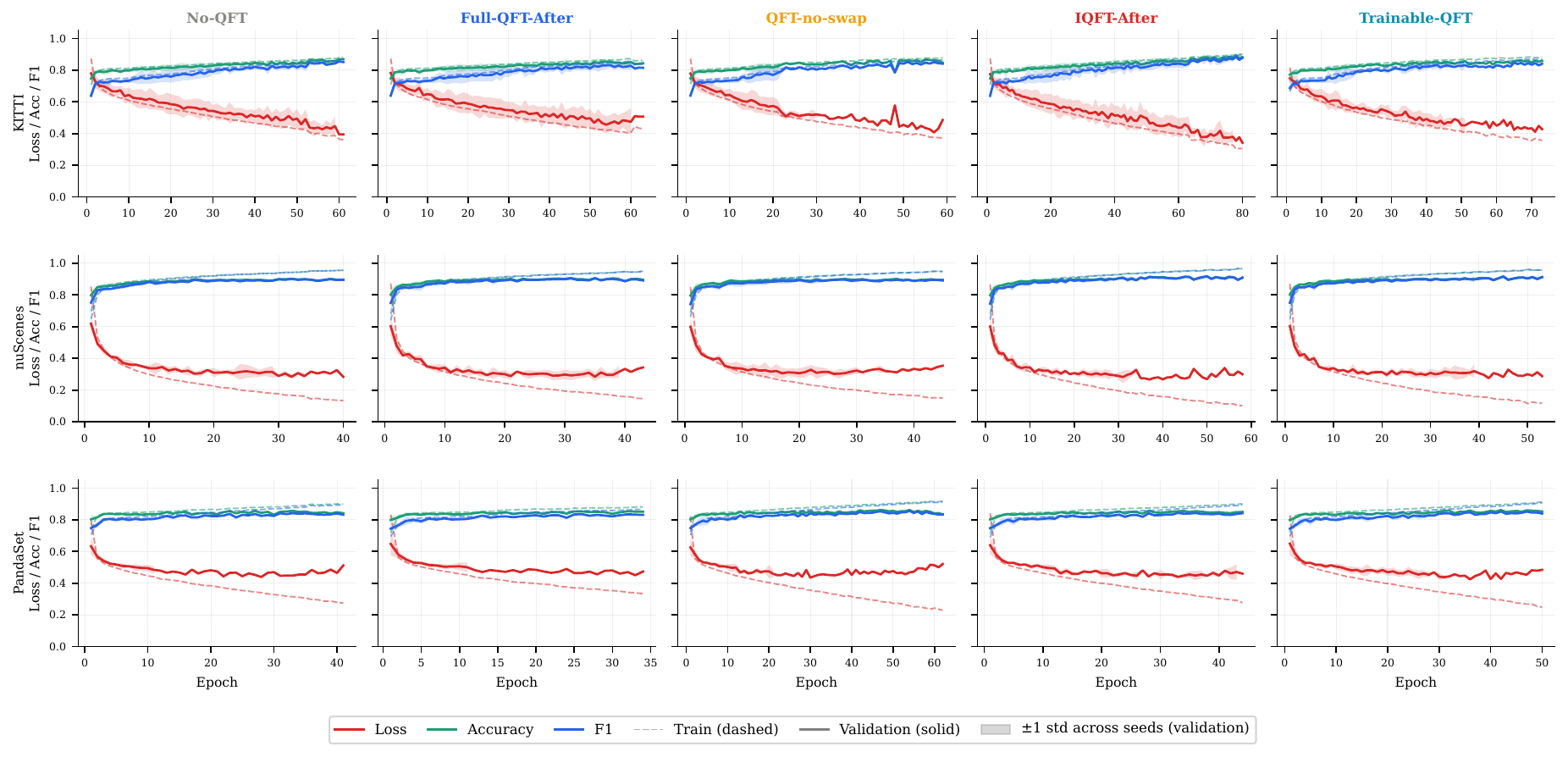}
    \caption{Training and validation dynamics for the five circuit
    configurations across KITTI, nuScenes, and PandaSet
    (rows: datasets; columns: configurations).
    Each panel presents loss, accuracy, and weighted F1,
    with solid and dashed curves denoting validation
    and training metrics, respectively. Shaded bands show $\pm1$ standard
    deviation of the validation metrics across the three seeds, drawn where at
    least two seeds are still training (seeds stop at different epochs under
    early stopping).}
    \label{fig:train_val_combined}
\end{figure*}

Figure~\ref{fig:train_val_combined} presents the evolution of
training and validation loss, accuracy, and weighted F1 for the
five configurations across the three datasets. These trajectories
provide an additional perspective on optimization behavior
and the development of classification performance during training.

On KITTI, the convergence profiles exhibit more pronounced
differences among configurations, consistent with the variation
in their final validation scores. IQFT-After achieves the highest
mean validation F1, while QFT-no-swap exhibits greater variability
across training runs. On nuScenes, the configurations reach
similar final validation performance, indicating convergence
toward comparable classification outcomes. PandaSet shows
moderate differences in final performance, with Trainable-QFT
obtaining the highest mean validation F1.

The comparison highlights two aspects of Fourier processing:
its influence on the final classification performance varies
across datasets, and additional Fourier operations do not
consistently improve the learned classifier. 

\subsection{Per-Class Classification Performance}
\label{subsec:perclass_performance}

\begin{figure*}[ht]
    \centering
    \includegraphics[width=\linewidth]
    {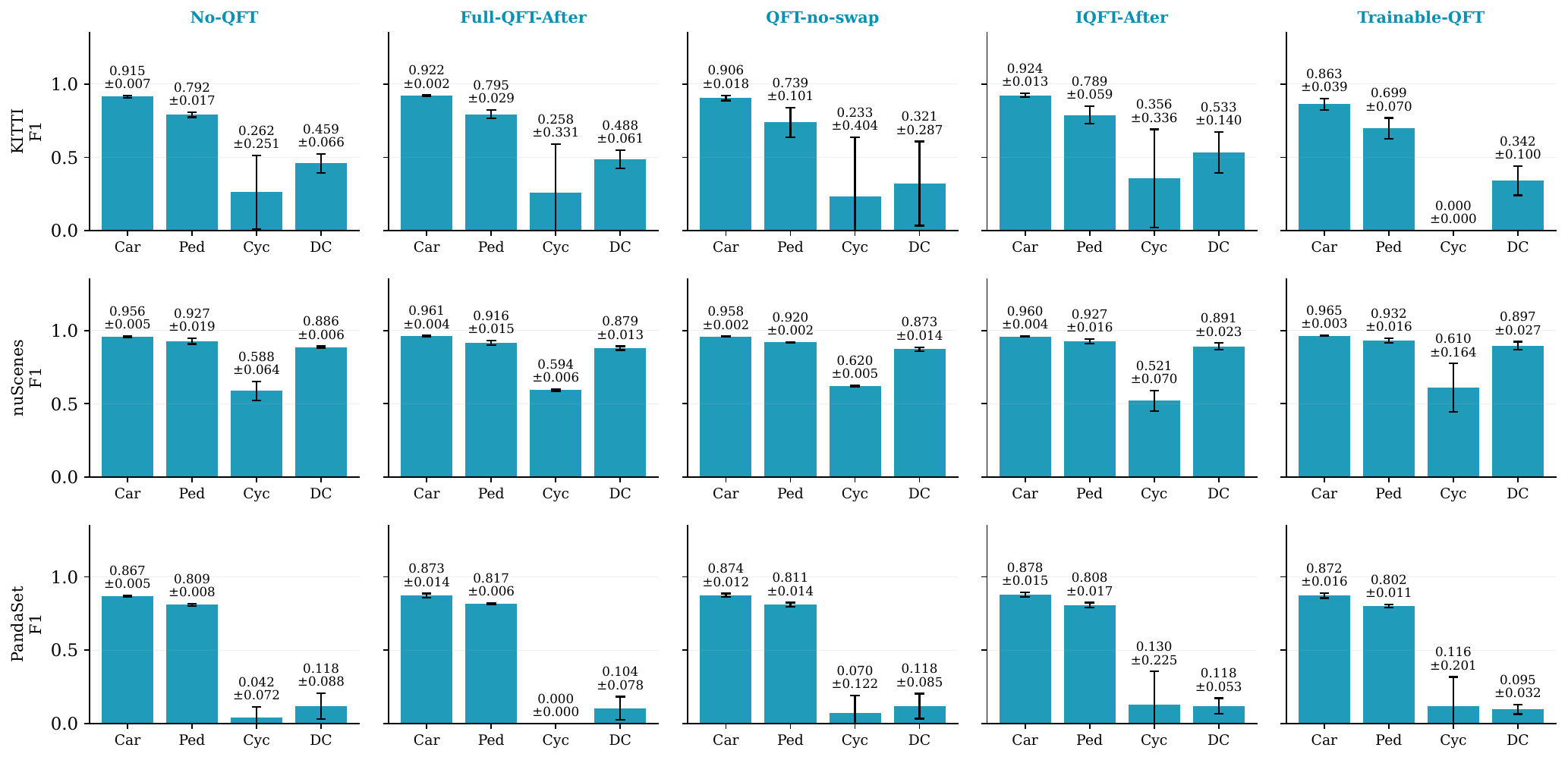}
    \caption{Per-class F1 scores of the five circuit
    configurations across KITTI, nuScenes, and PandaSet
    (rows: datasets; columns: circuit configurations).
    Bars represent the mean F1 score of each object
    category across three random seeds, with error
    bars indicating standard deviation. Numerical
    values are displayed above the bars.}
    \label{fig:percircuit}
\end{figure*}

The aggregate classification results are further examined
at the class level to characterize the influence of
Fourier-processing configurations on individual object
categories. As shown in Fig.~\ref{fig:percircuit},
the per-class F1 scores reveal dataset-dependent
differences in classification performance across
the five circuit configurations.

On KITTI, Cyclist exhibits pronounced variation across
configurations, with IQFT-After achieving the highest
F1 ($0.356$), followed by No-QFT ($0.262$) and
Full-QFT-After ($0.258$). Trainable-QFT records an
F1 of $0.000$ for this category. Car classification
remains more consistent, with F1 scores ranging from
$0.863$ to $0.924$ across configurations. These
results indicate that Fourier-processing choices
influence class-specific recognition, with greater
variation observed for the less frequently represented
Cyclist category.

On nuScenes, Cyclist F1 ranges from $0.521$ to
$0.620$, with QFT-no-swap achieving the highest
score. The narrower performance range indicates
more consistent Cyclist recognition across
configurations compared with KITTI. PandaSet
exhibits a different pattern, with IQFT-After
achieving the highest Cyclist F1 ($0.130$),
followed by Trainable-QFT ($0.116$) and
QFT-no-swap ($0.070$). Full-QFT-After records
an F1 of $0.000$ for this category. Across
PandaSet configurations, Car and Pedestrian
recognition remains relatively stable, with
F1 scores ranging from $0.867$ to $0.878$
and $0.802$ to $0.817$, respectively.

These findings characterize the relationship
between aggregate classification performance and
class-specific recognition. Configurations with
similar weighted F1 scores exhibit distinct
performance across individual categories,
particularly Cyclist. The variation across
KITTI, nuScenes, and PandaSet further demonstrates
that the effectiveness of Fourier processing
depends on the dataset and target class,
providing additional insight into the aggregate
rankings reported in Table~\ref{tab:simulated_3ds}.

\subsection{Representation Quality and Circuit Resource Cost}
\label{subsec:representation_resources}

The quality of the learned quantum representations and their
associated circuit costs are examined to assess the relationship
between discriminative performance and computational resources.
Representation quality is evaluated using four metrics:
silhouette score, Davies--Bouldin (DB) index,
inter/intra-class distance ratio, and centered kernel target
alignment (CKTA). The metrics are computed from the
quantum output feature vectors, with results averaged over
three training seeds. Circuit efficiency is assessed through
logical operations and backend-targeted transpilation.

\textit{Representation quality.}
Table~\ref{tab:repr_quality} summarizes the representation
metrics across the three datasets. On KITTI, IQFT-After
achieves the highest silhouette score ($0.197\pm0.047$),
inter/intra-class ratio ($1.313\pm0.385$), and CKTA
($0.444\pm0.027$), with the lowest DB index
($1.649\pm0.316$). This consistent ranking aligns with
its leading mean classification performance on KITTI,
indicating that improved feature separation accompanies
higher predictive performance for this configuration.

The relationship is less uniform on nuScenes and PandaSet.
On nuScenes, Full-QFT-After achieves the highest silhouette
score, Trainable-QFT leads in DB index and inter/intra-class
ratio, and IQFT-After obtains the highest CKTA. On PandaSet,
Full-QFT-After leads three representation metrics, despite
Trainable-QFT achieving the highest weighted classification
F1. This distinction indicates that favorable clustering
geometry does not consistently translate into the highest
classification performance. The observed variation across
seeds further highlights the sensitivity of representation
quality to the learned circuit parameters.

\begin{table*}[ht]
\centering
\caption{Quantum output representation quality across KITTI,
nuScenes, and PandaSet (mean $\pm$ standard deviation over
three seeds). Higher values are preferred for silhouette
score, inter/intra-class ratio, and CKTA; lower values
are preferred for DB index. Best mean values per dataset
are highlighted in bold.}
\label{tab:repr_quality}
\renewcommand{\arraystretch}{1.08}
\setlength{\tabcolsep}{6pt}
\small
\begin{tabular}{clcccc}
\toprule
& \textbf{Configuration}
& \textbf{Silhouette $\uparrow$}
& \textbf{DB $\downarrow$}
& \textbf{Inter/Intra $\uparrow$}
& \textbf{CKTA $\uparrow$} \\
\midrule

\multirow{5}{*}{\rotatebox[origin=c]{90}{\textit{KITTI}}}
& No-QFT
& $0.169\pm0.025$ & $1.960\pm0.499$
& $0.996\pm0.066$ & $0.407\pm0.021$ \\
& Full-QFT-After
& $0.174\pm0.031$ & $1.772\pm0.396$
& $1.175\pm0.122$ & $0.417\pm0.026$ \\
& QFT-no-swap
& $0.127\pm0.116$ & $2.746\pm1.510$
& $0.900\pm0.335$ & $0.363\pm0.085$ \\
& IQFT-After
& $\mathbf{0.197\pm0.047}$
& $\mathbf{1.649\pm0.316}$
& $\mathbf{1.313\pm0.385}$
& $\mathbf{0.444\pm0.027}$ \\
& Trainable-QFT
& $0.120\pm0.040$ & $2.213\pm0.217$
& $0.698\pm0.143$ & $0.326\pm0.070$ \\

\midrule

\multirow{5}{*}{\rotatebox[origin=c]{90}{\textit{nuScenes}}}
& No-QFT
& $0.436\pm0.030$ & $1.013\pm0.038$
& $2.454\pm0.191$ & $0.752\pm0.013$ \\
& Full-QFT-After
& $\mathbf{0.475\pm0.002}$
& $0.925\pm0.026$
& $3.028\pm0.149$ & $0.764\pm0.010$ \\
& QFT-no-swap
& $0.458\pm0.039$ & $0.962\pm0.071$
& $2.986\pm0.554$ & $0.764\pm0.008$ \\
& IQFT-After
& $0.465\pm0.021$ & $0.936\pm0.130$
& $2.967\pm0.551$
& $\mathbf{0.778\pm0.018}$ \\
& Trainable-QFT
& $0.473\pm0.033$
& $\mathbf{0.893\pm0.112}$
& $\mathbf{3.289\pm0.640}$
& $0.771\pm0.012$ \\

\midrule

\multirow{5}{*}{\rotatebox[origin=c]{90}{\textit{PandaSet}}}
& No-QFT
& $0.296\pm0.101$ & $3.419\pm0.770$
& $0.838\pm0.163$ & $0.622\pm0.016$ \\
& Full-QFT-After
& $\mathbf{0.317\pm0.086}$
& $\mathbf{3.123\pm0.451}$
& $0.892\pm0.085$
& $\mathbf{0.634\pm0.002}$ \\
& QFT-no-swap
& $0.292\pm0.038$ & $3.320\pm0.971$
& $0.900\pm0.119$ & $0.628\pm0.011$ \\
& IQFT-After
& $0.261\pm0.077$ & $3.299\pm1.142$
& $\mathbf{1.000\pm0.050}$
& $0.633\pm0.015$ \\
& Trainable-QFT
& $0.228\pm0.039$ & $3.273\pm1.673$
& $0.989\pm0.101$ & $0.628\pm0.018$ \\

\bottomrule
\end{tabular}
\end{table*}

\textit{Circuit resource cost.}
Table~\ref{tab:circuit_resources} compares the logical
two-qubit operation counts with the native CNOT counts
and circuit depths obtained after transpilation to IBM's
FakeManilaV2 target. The analysis considers the
variational and Fourier-processing stages, isolating
the additional circuit cost introduced by each
Fourier configuration.

No-QFT has the lowest resource requirements, with
16 logical two-qubit operations and 44 transpiled
CNOT gates at optimization level 3. Incorporating
the full QFT increases these counts to 24 logical
two-qubit operations and 68 compiled CNOT gates.
The corresponding depth increases from 90 to 130,
reflecting the additional Fourier-processing
operations and backend compilation overhead.

Removing the final QFT SWAP gates reduces the logical
two-qubit count from 24 to 22. At optimization level 0,
this simplification reduces the transpiled CNOT count
from 91 to 85, corresponding to a reduction of
approximately 6.6\%. At optimization level 3,
Full-QFT-After and QFT-no-swap both require
68 CNOT gates and achieve depth 130. The logical
reduction in gate count consequently translates into
different compiled resource savings depending on
the optimization level.

IQFT-After requires the same number of logical
two-qubit operations as Full-QFT-After and compiles
to 68 CNOT gates at optimization level 3, with
a depth of 129. Trainable-QFT retains the Fourier
circuit topology while increasing the number of
trainable quantum parameters from 48 to 54.
Its resource requirements are comparable to
Full-QFT-After under the representative phase
settings used for compilation.

\begin{table*}[ht]
\centering
\caption{Logical and FakeManilaV2-targeted quantum circuit
resources, excluding amplitude-state preparation.
O0 and O3 denote Qiskit optimization levels 0 and 3.
Two-qubit gate counts after transpilation correspond
to native CNOT operations. Quantum parameters refer
to trainable circuit parameters, excluding classical
model parameters.}
\label{tab:circuit_resources}
\renewcommand{\arraystretch}{1.12}
\setlength{\tabcolsep}{7pt}
\small
\begin{tabular}{lccccc}
\toprule
\textbf{Configuration}
& \textbf{Logical 2Q}
& \textbf{CNOT (O0)}
& \textbf{CNOT (O3)}
& \textbf{Depth (O3)}
& \textbf{Quantum Parameters} \\
\midrule
No-QFT
& \textbf{16} & \textbf{55} & \textbf{44}
& \textbf{90} & 48 \\
Full-QFT-After
& 24 & 91 & 68 & 130 & 48 \\
QFT-no-swap
& 22 & 85 & 68 & 130 & 48 \\
IQFT-After
& 24 & 82 & 68 & 129 & 48 \\
Trainable-QFT
& 24 & 91 & 68 & 130 & 54 \\
\bottomrule
\end{tabular}
\end{table*}

The circuit analysis further accounts for
amplitude-state preparation. For representative
amplitude-encoded inputs, including state preparation
increases the optimization-level-3 CNOT count from
68 to 89 for Full-QFT-After and from 44 to 70 for
No-QFT. These increases highlight the contribution
of state preparation to the full quantum execution
cost, alongside the variational and Fourier stages.

The results establish distinct relationships between
representation quality, classification performance,
and circuit resources. IQFT-After exhibits favorable
feature geometry and the highest mean weighted F1
on KITTI, while No-QFT consistently requires fewer
quantum operations. The comparison between
Full-QFT-After and QFT-no-swap further demonstrates
that logical circuit simplification does not
necessarily produce equivalent reductions after
backend-specific optimization. These findings
motivate evaluating Fourier-processing configurations
through both their learned representations and
their compiled circuit requirements.


\subsection{Cross-Dataset Performance and Hardware Robustness}\label{sec:generalisation}

The performance and hardware robustness of the five
Fourier-processing configurations are examined across KITTI,
nuScenes, and PandaSet. The cross-dataset evaluation assesses
the consistency of classification performance under different
LiDAR data distributions, while hardware robustness is
investigated using IBM's \texttt{FakeManilaV2} backend, which
provides a hardware-calibrated simulation environment
incorporating device-specific gate and measurement errors.

\textit{Cross-dataset performance.}
The classification results in Table~\ref{tab:simulated_3ds}
demonstrate variation in the relative effectiveness of
Fourier processing across the three datasets. IQFT-After
achieves the highest mean weighted F1 on KITTI
($0.857\pm0.033$), whereas Full-QFT-After and IQFT-After
share the highest mean on nuScenes ($0.906$). On PandaSet,
Trainable-QFT achieves the strongest performance
($0.836\pm0.018$). The variation in circuit rankings
indicates that the benefits of Fourier processing depend
on the characteristics of the input data and learned
representations.

The performance differences are particularly limited
on nuScenes, where the weighted F1 scores of all five
configurations fall within $0.901$--$0.906$. In contrast,
KITTI exhibits a wider performance range
($0.803$--$0.857$), accompanied by greater variability
across training seeds for selected configurations.
PandaSet presents an intermediate range
($0.816$--$0.836$), with Trainable-QFT achieving the
highest mean classification performance. These results
characterize the consistency of circuit behavior across
datasets and indicate that the relative advantages of
individual Fourier configurations are data-dependent.

\textit{Hardware robustness.}
The trained models are further evaluated under
hardware-calibrated simulation using
\texttt{FakeManilaV2}. Table~\ref{tab:hardware_col}
reports the weighted F1 under ideal and hardware-calibrated
execution, with the performance difference
\begin{equation}
\Delta_{\mathrm{F1}}
=
\mathrm{F1}_{\mathrm{ideal}}
-
\mathrm{F1}_{\mathrm{hardware}}.
\label{eq:f1_hardware_drop}
\end{equation}
Both execution settings use identical evaluation samples
and trained checkpoints, enabling direct assessment of
hardware-induced performance degradation.

On KITTI, Trainable-QFT achieves the highest
hardware-calibrated F1 ($0.793\pm0.041$) and the
smallest degradation ($\Delta_{\mathrm{F1}}=0.053$).
Full-QFT-After follows with an F1 of
$0.784\pm0.040$ and a degradation of $0.067$.
Although IQFT-After achieves the highest ideal F1
($0.863\pm0.036$), its performance decreases to
$0.755\pm0.006$ under hardware-calibrated execution.
QFT-no-swap exhibits the largest variability,
with an F1 of $0.677\pm0.173$.

On nuScenes, No-QFT achieves the highest
hardware-calibrated F1 ($0.882\pm0.018$) and
the smallest degradation ($0.040$).
IQFT-After experiences the largest reduction,
from $0.930\pm0.010$ under ideal simulation
to $0.710\pm0.112$ under hardware-calibrated
execution. PandaSet exhibits a similar trend,
with No-QFT retaining the highest F1
($0.799\pm0.030$) and the smallest degradation
($0.017$). The Fourier-based configurations
experience larger reductions, ranging from
$0.159$ for Full-QFT-After to $0.239$ for
QFT-no-swap.

\begin{table}[ht]
\centering
\caption{Weighted F1 under ideal and FakeManilaV2
hardware-calibrated simulation (mean $\pm$ standard
deviation over three trained checkpoints).
$\Delta_{\mathrm{F1}}$ denotes the difference between ideal and hardware-calibrated F1. Ideal F1
is computed on the same $\sim$200-sample evaluation subset as the hardware F1
and therefore differs from Table~\ref{tab:simulated_3ds}.}
\label{tab:hardware_col}
\renewcommand{\arraystretch}{1.12}
\setlength{\tabcolsep}{3pt}
\scriptsize
\begin{tabular}{clccc}
\toprule
& \textbf{Configuration}
& \textbf{Ideal F1}
& \textbf{Hardware F1}
& $\boldsymbol{\Delta_{\mathrm{F1}}}$ \\
\midrule

\multirow{5}{*}{\rotatebox[origin=c]{90}{\textit{KITTI}}}
& No-QFT
& $0.842\pm0.028$ & $0.756\pm0.027$ & $0.086$ \\
& Full-QFT-After
& $0.851\pm0.013$ & $0.784\pm0.040$ & $0.067$ \\
& QFT-no-swap
& $0.814\pm0.046$ & $0.677\pm0.173$ & $0.137$ \\
& IQFT-After
& $0.863\pm0.036$ & $0.755\pm0.006$ & $0.108$ \\
& Trainable-QFT
& $0.847\pm0.033$
& $\mathbf{0.793\pm0.041}$
& $\mathbf{0.053}$ \\

\midrule

\multirow{5}{*}{\rotatebox[origin=c]{90}{\textit{nuScenes}}}
& No-QFT
& $0.922\pm0.023$
& $\mathbf{0.882\pm0.018}$
& $\mathbf{0.040}$ \\
& Full-QFT-After
& $0.908\pm0.024$ & $0.733\pm0.102$ & $0.174$ \\
& QFT-no-swap
& $0.908\pm0.015$ & $0.722\pm0.011$ & $0.185$ \\
& IQFT-After
& $0.930\pm0.010$ & $0.710\pm0.112$ & $0.220$ \\
& Trainable-QFT
& $0.919\pm0.009$ & $0.774\pm0.141$ & $0.144$ \\

\midrule

\multirow{5}{*}{\rotatebox[origin=c]{90}{\textit{PandaSet}}}
& No-QFT
& $0.817\pm0.034$
& $\mathbf{0.799\pm0.030}$
& $\mathbf{0.017}$ \\
& Full-QFT-After
& $0.806\pm0.007$ & $0.647\pm0.227$ & $0.159$ \\
& QFT-no-swap
& $0.816\pm0.029$ & $0.577\pm0.098$ & $0.239$ \\
& IQFT-After
& $0.825\pm0.026$ & $0.631\pm0.110$ & $0.194$ \\
& Trainable-QFT
& $0.836\pm0.022$ & $0.673\pm0.193$ & $0.163$ \\

\bottomrule
\end{tabular}
\end{table}

\textit{Relationship with circuit resources.}
The hardware-calibrated results can be examined
alongside the backend-transpiled resource measurements
in Table~\ref{tab:circuit_resources}. No-QFT has
the lowest compiled two-qubit gate count and circuit
depth, consistent with its comparatively strong
hardware robustness on nuScenes and PandaSet.
However, configurations with similar compiled
resource requirements exhibit different performance
degradation. At optimization level 3,
Full-QFT-After, IQFT-After, and Trainable-QFT
each require 68 CNOT gates, yet their
hardware-calibrated F1 scores differ across
all three datasets.

This variation indicates that hardware robustness
is influenced by both circuit-resource requirements
and the specific quantum transformations and
learned parameters. A higher
classification score under ideal simulation
does not necessarily correspond to stronger
performance under hardware-calibrated execution.
Across the evaluated datasets, the results
highlight the importance of considering
cross-dataset performance consistency and
hardware robustness when selecting
Fourier-processing configurations for
hybrid quantum classifiers.

\subsection{Comparison with Existing Classical and Quantum Baselines}
\label{sec:baselines}

The proposed hybrid framework is compared with a
parameter-matched classical architecture and contextualized
against established and recent methods for LiDAR-based
perception. The analysis considers classification performance,
model complexity, and developments in classical and quantum
point-cloud processing. Table~\ref{tab:baselines}
summarizes the reported results across KITTI, nuScenes,
and PandaSet.

\textit{Comparison with a parameter-matched classical model.}
The classical baseline uses the same PointNet backbone, feature compression,
entropy descriptor, and classification head as the hybrid model, with the
variational quantum circuit replaced by a small classical network
($16\rightarrow8\rightarrow4$ with tanh activation) that produces a
four-dimensional output of the same shape as the Pauli-$Z$ expectation vector.
It contains 45,908 trainable parameters, compared with 45,784--45,790 for the
hybrid architectures, and is trained on the same data, seeds, and early-stopping
protocol, with an 80/20 train--validation split per seed. This comparison evaluates the effect of replacing the classical feature-processing module with the quantum circuit under matched training conditions.

On KITTI, the classical model achieves $82.92 \pm 2.08$\%
accuracy, below all five hybrid configurations. IQFT-After
achieves $87.30 \pm 2.50$\%, an improvement of 4.38 percentage
points, with classification Car AP increasing from
$89.5 \pm 5.8$ to $96.2 \pm 1.0$. On nuScenes, IQFT-After reaches
$90.80\pm0.80\%$ compared with $89.56\pm0.51\%$ for the classical model
(+1.2 points), and its mean AP over the three object classes is higher
($85.3\pm3.8$ versus $72.8\pm2.9$), with Cyclist AP increasing from $23.6$ to $59.4$. On PandaSet, Trainable-QFT achieves $85.20 \pm 1.10$\%
accuracy, compared with $85.31 \pm 0.44$\% for the classical
MLP. The hybrid advantage is
therefore dataset-dependent: clearest on KITTI, smaller on nuScenes, and absent
on PandaSet. 
\begin{table*}[!t]
\centering
\caption{Comparison of existing LiDAR perception methods and the proposed hybrid framework on KITTI, nuScenes, and PandaSet. Detection methods are reported using AP or mAP under their respective evaluation protocols; our classifiers are reported using validation accuracy and classification AP (Cls.\ AP), as mean $\pm$ standard deviation over three seeds. Cls.\ AP is computed from softmax scores on a fixed class-stratified subset of 500 samples per dataset, using full scans for KITTI and ground-truth-cropped objects for nuScenes and PandaSet. Cls.\ AP corresponds to Car AP on KITTI and PandaSet and mean AP over Car, Pedestrian, and Cyclist on nuScenes. The classical MLP shares the backbone, compression layer, entropy descriptor, and classification head with the hybrid model, replacing only the quantum circuit with a small classical network. $\sim$ denotes an approximate or configuration-dependent value; -- denotes an unavailable or inapplicable entry. C, L, and R denote camera, LiDAR, and radar inputs, respectively. }
\label{tab:baselines}
\renewcommand{\arraystretch}{1.10}
\setlength{\tabcolsep}{5pt}
\small
\begin{tabular}{@{}cllllr@{}}
\toprule
& \textbf{Architecture}
& \textbf{Year}
& \textbf{Input}
& \textbf{Performance}
& \textbf{Parameters} \\
\midrule

\multirow{5}{*}{\rotatebox[origin=c]{90}{\textit{KITTI}}}
& PointPillars~\cite{lang2019pointpillars}
& 2019 & L
& 82.58 AP (Car, Easy, 3D)
& $\sim$4.8M \\

& Kulkarni~\cite{kulkarni2025lidar}
& 2025 & C+L
& 87.0\% val.\ acc.
& $\sim$1M \\

& ImagePG~\cite{lee2026image}
& 2026 & C+L
& 82.46 AP (Car, Mod., 3D)
& 52.54M \\

& Classical MLP (ours)
& -- & L
& \shortstack[l]{
    $82.92 \pm 2.08$\% val.\ acc.\\
    $89.5 \pm 5.8$ subset Cls.\ AP}
& 45,908 \\

& \textbf{IQFT-After (ours)}
& -- & L
& \shortstack[l]{
    $87.30 \pm 2.50$\% val.\ acc.\\
    $96.2 \pm 1.0$ subset Cls.\ AP}
& 45,784 \\

\midrule

\multirow{7}{*}{\rotatebox[origin=c]{90}{\textit{nuScenes}}}
& PointPillars~\cite{lang2019pointpillars}
& 2019 & L
& --
& $\sim$4.8M \\

& TransFusion-L~\cite{bai2022transfusion}
& 2022 & L
& 65.52 mAP
& $\sim$40M \\

& SEGT~\cite{segt2024}
& 2024 & L
& 70.1 mAP (test, no test-time augmentation)
& $\sim$50M \\

& FDSNet~\cite{mohammed2025fdsnet}
& 2025 & C+L+R
& 75.9 mAP (val.)
& 82.4M \\

& BEVDilation~\cite{zhang2026bevdilation}
& 2026 & C+L
& 73.1 mAP (test)
& -- \\

& Classical MLP (ours)
& -- & L
& \shortstack[l]{
    $89.56 \pm 0.51$\% val.\ acc.\\
    $72.8 \pm 2.9$ subset Cls.\ mAP}
& 45,908 \\

& \textbf{IQFT-After (ours)}
& -- & L
& \shortstack[l]{
    $90.80 \pm 0.80$\% val.\ acc.\\
    $85.3 \pm 3.8$ subset Cls.\ mAP}
& 45,784 \\

\midrule

\multirow{5}{*}{\rotatebox[origin=c]{90}{\textit{PandaSet}}}
& PV-RCNN~\cite{xiao2021pandaset}
& 2021 & L
& 79.56 AP (Car, 0--50\,m)
& $\sim$12M \\

& CenterPoint~\cite{yin2021center}
& 2021 & L
& 78.38 AP (IoU 0.7, Mod.)
& $\sim$21.8M \\

& CPD~\cite{wu2024commonsense}
& 2024 & L
& 55.07 AP (IoU 0.7, Mod.)
& -- \\

& Classical MLP (ours)
& -- & L
& \shortstack[l]{
    $85.31 \pm 0.44$\% val.\ acc.\\
    $95.6 \pm 1.3$ subset Cls.\ AP}
& 45,908 \\

& \textbf{Trainable-QFT (ours)}
& -- & L
& \shortstack[l]{
    $85.20 \pm 1.10$\% val.\ acc.\\
    $95.5 \pm 0.8$ subset Cls.\ AP}
& 45,790 \\

\bottomrule
\end{tabular}
\end{table*}

\textit{Comparison with classical LiDAR architectures.}
Classical LiDAR perception has progressed from
pillar-based representations, such as PointPillars
\cite{lang2019pointpillars}, toward architectures
incorporating sparse voxel processing, attention
mechanisms, and multimodal feature integration.
On KITTI, ImagePG~\cite{lee2026image}
uses image-guided pseudo-LiDAR point generation
to improve three-dimensional object detection.
The earlier multimodal PointNet framework
\cite{kulkarni2025lidar} provides an additional
reference for four-class classification using
RGB and LiDAR information.

On nuScenes, TransFusion-L~\cite{bai2022transfusion}
employs transformer-based detection with LiDAR
features, while SEGT~\cite{segt2024} uses spatial
expansion and group attention for point-cloud
representation learning. Recent multimodal
approaches, including BEVDilation
\cite{zhang2026bevdilation} and FDSNet
\cite{mohammed2025fdsnet}, further explore feature
integration across LiDAR, camera, and radar
sensors. For PandaSet, PV-RCNN provides a
supervised detection reference
\cite{xiao2021pandaset}, while CPD
\cite{wu2024commonsense} investigates
unsupervised three-dimensional object detection
using prototype-based pseudo-label refinement.

These architectures establish the development
of LiDAR-based perception across established
and recent methods. Their reported AP and mAP scores characterize three-dimensional
object detection under dataset-specific
protocols. Our framework addresses
four-class point-cloud classification, for
which the parameter-matched MLP provides
the direct quantitative comparison.

\textit{Comparison with quantum point-cloud methods.}
Quantum approaches to three-dimensional data
processing have explored representation
learning, permutation-invariant encoding,
and hybrid quantum--classical architectures.
Rathi et al.~\cite{rathi2023qae}
investigate quantum autoencoding for
three-dimensional point clouds, while
Heredge et al.~\cite{heredge2024perminv}
develop permutation-invariant quantum
representations. Hybrid quantum
convolutional architectures have also
been investigated for LiDAR-based
perception~\cite{roh2025hqcod}.

This study examines Fourier processing
within a common hybrid PointNet--quantum
architecture, comparing five circuit
configurations across three LiDAR datasets.
The evaluation considers aggregate and
class-specific classification performance,
quantum output representation quality,
hardware-calibrated robustness, and
logical and backend-transpiled circuit
resources. These dimensions characterize
the predictive behavior and implementation
cost of alternative quantum
Fourier-processing strategies.

\subsection{Discussion}
\label{subsec:discussion}

The findings highlight the dataset-dependent influence of Fourier processing
on hybrid quantum--classical LiDAR classification. IQFT-After achieves the
highest mean classification performance and consistently favorable
representation-quality metrics on KITTI, while the relative rankings vary
across nuScenes and PandaSet. This variation indicates that improved
feature separation does not consistently translate into higher predictive
performance, as the effectiveness of the quantum transformation depends
on the learned representations and classification task. The differences
in Cyclist recognition further demonstrate how aggregate weighted F1 can
conceal substantial variations in minority-class performance, emphasizing
the importance of class-specific evaluation and multi-seed comparisons.

The circuit analysis reveals a distinction between logical simplification
and backend-targeted efficiency. Removing the final SWAP gates in the QFT
circuit induces a permutation of the Pauli-$Z$ expectation values, which
can be absorbed into the subsequent linear classification layer without
reducing the model's expressive capacity under ideal execution. However,
the reduction in logical two-qubit operations does not consistently
translate into lower compiled resource requirements. At optimization
level 3, QFT-no-swap and Full-QFT-After exhibit identical CNOT counts
and circuit depths for the evaluated FakeManilaV2 target. This finding
highlights the importance of assessing circuit efficiency after
hardware-aware compilation, since logical gate reductions alone do
not fully characterize execution cost.

Hardware-calibrated evaluation further demonstrates that classification
performance under ideal simulation is not a reliable indicator of
hardware robustness. Trainable-QFT achieves the highest
hardware-calibrated F1 on KITTI, while No-QFT leads on nuScenes and
PandaSet. Moreover, configurations with comparable compiled gate counts
exhibit different performance degradation, indicating that circuit
complexity alone does not determine sensitivity to device-derived errors.
The results emphasize the interplay between circuit structure, learned
parameters, and hardware characteristics, motivating joint consideration
of predictive performance, resource efficiency, and hardware robustness.
The parameter-matched classical comparison reinforces this perspective,
with the relative benefit of quantum feature processing varying across the
evaluated datasets: the hybrid models lead on KITTI and nuScenes, while the models achieve similar mean accuracy on PandaSet.

\section{Conclusion}
\label{sec:conclusion}

This work presented a systematic evaluation of QFT processing within a hybrid PointNet--quantum neural network for LiDAR point-cloud classification. Five circuit configurations were
investigated across KITTI, nuScenes, and PandaSet using a common
architecture and multi-seed experimental protocol. The evaluation
examined classification accuracy, representation quality, logical and backend-transpiled circuit resources, and hardware-calibrated robustness.
The findings establish dataset-dependent differences in the effectiveness of Fourier processing, with IQFT-After achieving the highest mean weighted F1 on KITTI, IQFT-After and Full-QFT-After sharing the highest mean on nuScenes, and Trainable-QFT leading on PandaSet.

The resource analysis demonstrates that removing the final QFT SWAP gates reduces logical circuit complexity, while the benefits after transpilation depend on compiler optimization. Hardware-calibrated evaluation using IBM's FakeManilaV2 backend further reveals that ideal-simulation performance does not consistently predict robustness under device-derived errors. These findings establish a basis for evaluating Fourier-based hybrid quantum classifiers through predictive performance and hardware-oriented resource characteristics. Future work will investigate noise-aware Fourier processing, connectivity-aware circuit optimization, alternative quantum measurement strategies, and validation on physical quantum processors and larger point-cloud benchmarks.

\section*{Acknowledgment}
This work was supported in part by the NYUAD Center for Quantum and Topological Systems (CQTS), funded by Tamkeen under the NYUAD Research Institute grant CG008.
\section*{Generative AI Use Disclosure}

During the preparation of this work, the authors used Generative AI tools (specifically ChatGPT and Grammarly) for language editing and text refinement. The authors reviewed and edited all generated or refined content as needed and take full responsibility for the publication's content.

\bibliographystyle{IEEEtran}
\bibliography{refs}

\end{document}